%% file: main.tex
\documentclass[10pt,twocolumn,letterpaper]{article}

\usepackage{cvpr}              
\usepackage{cuted}


\definecolor{cvprblue}{rgb}{0.21,0.49,0.74}
\usepackage[pagebackref,breaklinks,colorlinks,allcolors=cvprblue]{hyperref}
\usepackage{multirow}

\def\paperID{} 
\def\confName{CVPR}
\def\confYear{2025}

\title{GS-DiT: Advancing Video Generation with Pseudo 4D Gaussian Fields through Efficient Dense 3D Point Tracking}

\author{
Weikang Bian$^{1,2}$\\
Yijin Li$^{3}$
\and
Zhaoyang Huang$^{3}$\\
Fu-Yun Wang$^{1}$
\and
Xiaoyu Shi$^{1}$\\
Hongsheng Li$^{1,2}$
\and
$^{1}$Multimedia Laboratory, The Chinese University of Hong Kong\\
$^{2}$Centre for Perceptual and Interactive Intelligence 
$^{3}$Avolution AI
}

\begin{document}
\maketitle

\begin{strip}\centering
    \centering
    \includegraphics[trim={2.5cm 1cm 1cm 1.5cm},clip,width=0.95\linewidth]{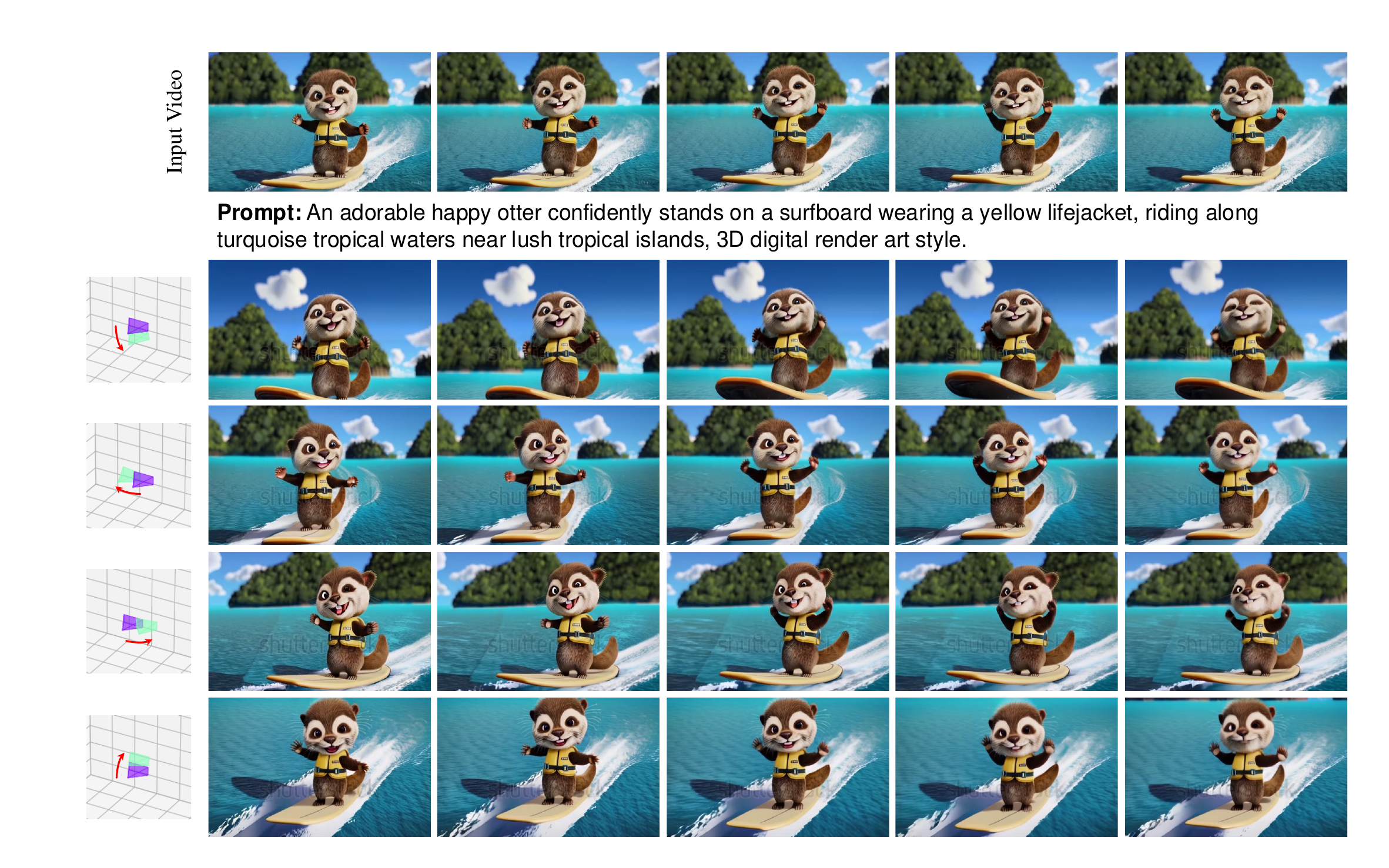}
\captionof{figure}{\label{fig:teaser}
GS-DiT generates multi-camera shooting videos by bringing pseudo 4D Gaussian fields to video diffusion transformers.
}
\end{strip}

\input{sec/0_abstract}    
\input{sec/1_intro}
\input{sec/2_relwk}
\input{sec/3_method}

\input{sec/4_exp}
\input{sec/5-conclusion}
{
    \small
    \bibliographystyle{ieeenat_fullname}
    \bibliography{main}
}

\clearpage
\setcounter{section}{0}
\setcounter{table}{0}
\setcounter{figure}{0}
\section*{Appendix}
\renewcommand\thesection{\Alph{section}}
\renewcommand{\thetable}{A\arabic{table}}
\renewcommand{\thefigure}{A\arabic{figure}}
\renewcommand{\theHtable}{\thesection\arabic{table}}

\input{sec/supp}


\end{document}

%% file: sec/0_abstract.tex
\begin{abstract}
4D video control is essential in video generation as it enables the use of sophisticated lens techniques, such as multi-camera shooting and dolly zoom, which are currently unsupported by existing methods. 
Training a video Diffusion Transformer (DiT) directly to control 4D content requires expensive multi-view videos.
Inspired by Monocular Dynamic novel View Synthesis (MDVS) that optimizes a 4D representation and renders videos according to different 4D elements, such as camera pose and object motion editing,
we bring pseudo 4D Gaussian fields to video generation.
Specifically, we propose a novel framework that constructs a pseudo 4D Gaussian field with dense 3D point tracking and renders the Gaussian field for all video frames.
Then we finetune a pretrained DiT to generate videos following the guidance of the rendered video, dubbed as GS-DiT.
To boost the training of the GS-DiT, we also propose an efficient Dense 3D Point Tracking (D3D-PT) method for the pseudo 4D Gaussian field construction.
Our D3D-PT outperforms SpatialTracker, the state-of-the-art sparse 3D point tracking method, in accuracy and accelerates the inference speed by two orders of magnitude.
During the inference stage, GS-DiT can generate videos with the same dynamic content while adhering to different camera parameters, addressing a significant limitation of current video generation models.
GS-DiT demonstrates strong generalization capabilities and extends the 4D controllability of Gaussian splatting to video generation beyond just camera poses. It supports advanced cinematic effects through the manipulation of the Gaussian field and camera intrinsics, making it a powerful tool for creative video production. Demos are available at \url{https://wkbian.github.io/Projects/GS-DiT/}.
\end{abstract}

%% file: sec/1_intro.tex
\section{Introduction}
\label{sec:intro}


%
%


Video generation~\cite{harvey2022flexible,shi2024motion,ho2022imagen,esser2023structure,videoworldsimulators2024,yang2024cogvideox,guo2023animatediff} has made rapid progress in recent years.
Since the demonstration of Sora~\cite{yang2024cogvideox}, the video generated by Diffusion Transformers (DiT) is approaching cinematic quality by scaling up the parameters and the number of training videos. 
However, in order to produce effective lens language, such as multi-camera shooting, dolly zoom, and object motion editing in videos, video creators often need to have more precise 4D control over the video content. For example, multi-camera shooting requires the presentation of the same content from different perspectives and dolly zoom requires adjusting the intrinsic and extrinsic parameters of the camera at the same time.
Yet, current video generation methods are unable to support such 4D video control.

Generative Camera Dolly~(GCD)~\cite{vanhoorick2024gcd} demonstrates the possibility of utilizing the generative model for multi-camera shooting.
After preparing multi-camera shooting videos as training data,
it switches videos captured by different camera trajectories and the same dynamic contents as input condition video and output supervision.
However, capturing a large amount of synchronized multi-trajectory videos in open-door real scenes is too expensive, \eg{}, car driving videos, so GCD chooses to collect training data in a simulator, which exhibits poor generalization performance to the real world and prohibits it from being trained on web videos.
This raises the question: 
can we directly learn multi-camera shooting video generation from normal monocular videos? 
One straightforward solution is to optimize a 4D Gaussian field~\cite{bansal20204d,wang2024shape,wu20244d} for each monocular video and then reprogram the camera trajectories to render novel view videos for training. 
However, the optimization process is highly time-consuming, requiring at least one hour for a 70-frame video, which makes the preparation of training data impractical. Furthermore, due to the ill-posed nature of 4D video optimization, the rendered videos suffer from significant artifacts. Fine-tuning video generation models with these videos will inevitably degrade the video quality.

Fortunately, we observe that the videos rendered from the Gaussian field although contain artifacts, still provide strong cues for video generation.
Drawing inspiration from 
recent works on
classical monocular dynamic video synthesis (MDVS)~\cite{gao2022monocular},
we build the pseudo 4D Gaussian field and render the novel view videos to guide the video generation.
Specifically, we first propose an efficient dense 3D point-tracking (D3D-PT) method to facilitate the pseudo 4D Gaussian field construction.
Given an input video, we select a reference frame and estimate its dense 3D point trajectories across the subsequent frames.
The pseudo 4D Gaussian field is then directly constructed from 3D point trajectories without any optimization. The Gaussian primitives~\cite{3d_gs} of the field have constant color values derived from the pixel values, replacing the view-dependent spherical harmonic representation. Other parameters of the Gaussian primitives, such as opacity, are all held constant.
We then generate the training data by rendering the Gaussian field using the original camera poses from the input video.
A pretrained Video DiT is finetuned with the rendered results as input to generate the original input video from the rendered video, dubbed as GS-DiT.
Since it is difficult to annotate the ground truth of 3D point tracking in the real world, point trackers are trained on synthetic data, and the depth distribution in the real world may be quite different from that in synthetic data. Our D3D-PT is loosely coupled to the depth distribution so that can be flexibly adapted to different scenarios.
In the inference stage, given a monocular dynamic video, we also build a pseudo 4D Gaussian field and render it according to the newly assigned camera parameters.
Then, our finetuned DiT generates the final video following the guidance of the rendered video.
Besides the camera pose control, we can also adjust the camera intrinsics and edit the Gaussian field.
Such 4D control of video generation supports video creators.

Our proposed framework does not require multi-view videos for training. GS-DiT can therefore be trained on any monocular dynamic videos and learn various motion dynamics.
Compared to GCD, GS-DiT presents a strong generalization capacity and can be consistently improved when the base DiT model is scaled up. 
GS-DiT can also support more 4D controllability than GCD, such as camera intrinsic control and object motion editing.
The main contributions of this paper are summarized as follows:



\begin{itemize}
    \item We propose a novel framework that advances video generation with pseudo 4D Gaussian fields through efficient dense 3D point tracking.
    \item We propose an efficient dense 3D point tracking (D3D-PT) method that outperforms SpatialTracker in accuracy, and accelerates dense 3D point tracking by two orders of magnitude. D3D-PT facilitates the training of video generation from a rendered pseudo 4D Gaussian field.
    \item We propose GS-DiT that learns to generate videos following the guidance of rendered videos. GS-DiT can generate multi-camera shooting videos and provides 4D video control such as camera intrinsic and object motion editing.
\end{itemize}

\begin{figure*}[t]
    \centering
    \includegraphics[trim={2cm 2cm 0.8cm 1.5cm},clip,width=0.95\textwidth]{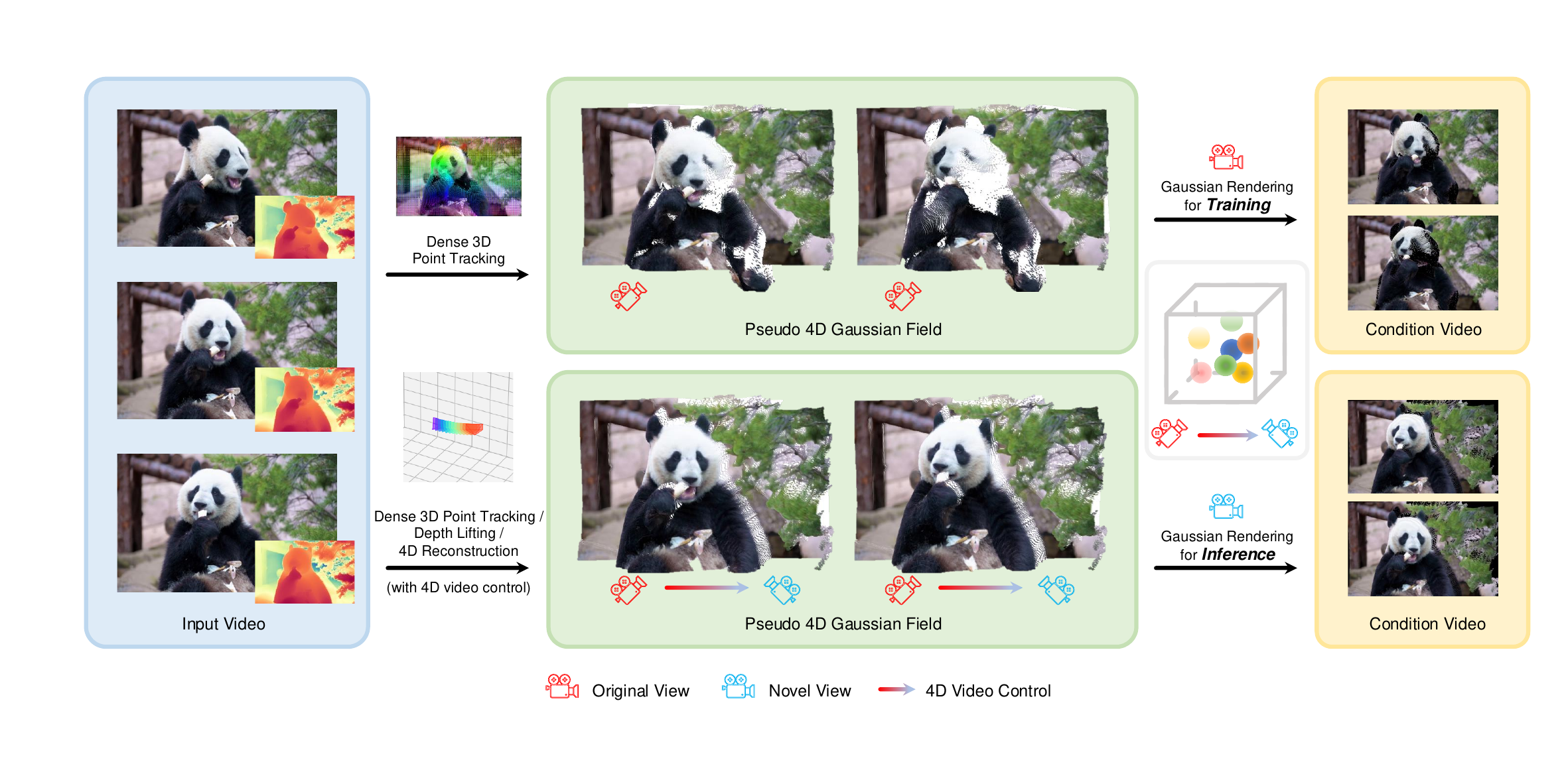}
    \vspace{-0.3cm}
    \caption{\textbf{An overview of GS-DiT.} In the training stage, we build a pseudo 4D Gaussian field from an input video via dense 3D point tracking. Our GS-DiT learns to generate the original video guided by the video rendered from the pseudo 4D Gaussian field. In the inference stage, we can build the pseudo 4D Gaussian via dense 3D point tracking, directly lifting the depth map, or optimizing a 4D Gaussian field. Editing and rendering the Gaussian field with scheduled camera intrinsic and extrinsic bring various cinematic effects. 
    }
    \vspace{-0.5cm}
    \label{fig:pseudo_4dgs}
\end{figure*}

%% file: sec/2_relwk.tex
\section{Related Work}
\label{sec:relwk}

\noindent \textbf{Point tracking.} PIPs~\cite{harley2022particle} and TAP-Net~\cite{doersch2022tap} first address the point-tracking problem that estimates point trajectories in videos for the query pixels in the starting video frame. In contrast to optical flow that tackles dense pixel correspondence between a pair of video frames, they are interested in independent pixel correspondence throughout multiple video frames.
CoTracker~\cite{karaev23cotracker} and Context-PIPs~\cite{bian2023contextpips} realize that even though focusing on long-term temporal pixel correspondence, spatial context information is still vital and improves point tracking accuracy with more spatial information.
DOT~\cite{lemoing2024dense} estimates dense point tracking by incorporating a pretrained point tracker to provide initialization and refine the dense prediction with a RAFT~\cite{teed2020raft}.
SpatialTracker~\cite{SpatialTracker} is the first method that extends the 2D point tracking to 3D via a depth estimator.
In our GS-DiT framework, we need a dense 3D point tracker to build the pseudo-4D Gaussian field but existing point trackers cannot complete this task.
We propose the first dense 3D point tracking method, surpassing SpatialTracker by accuracy and accelerating the speed by two orders.

\noindent \textbf{Video diffusion models.}
Diffusion models~\cite{ho2020denoising,dhariwal2021diffusion,nichol2021improved,liu2022flow,wang2024rectified,karras2022elucidating} emerge as a prominent approach for generative modeling, excelling in synthesizing diverse and high-quality samples. Although early diffusion models are primarily validated on image generation tasks~\cite{dhariwal2021diffusion,ho2022classifier}, their application is soon extended to video generation, revealing significant potential. 
Initial video diffusion models~\cite{ho2022video,blattmann2023align,harvey2022flexible,shi2024motion,ho2022imagen,esser2023structure} are often based on off-the-shelf image diffusion models~\cite{rombach2022high}, augmented with temporal layers to capture the relationships among frames. 
Since Sora~\cite{videoworldsimulators2024}, video generation models have begun to shift to the transformer architectures, such as CogVideoX series~\cite{yang2024cogvideox}.
Video Diffusion with Transformers~(DiT)~\cite{peebles2023scalable} utilizes temporal causal VAE~\cite{yu2023language} and 3D attention for better temporal consistency.
Recent works show that image and video diffusion models can be re-purposed for other dense video translation tasks including monocular depth estimation~\cite{shao2024learning,garcia2024fine,ke2024repurposing}, 3D generation and reconstruction~\cite{voleti2025sv3d,blattmann2023stable,chen2024v3d,poole2022dreamfusion}, and amodal segmentation~\cite{ozguroglu2024pix2gestalt}.
Text-to-4D, image-to-4D, and Video-to-4D~\cite{van2022revealing} papers have also attracted researcher's attention but the scenarios are highly limited, \eg{}, single objects or animals.
In addition to the generation quality, the generation controllability, such as the camera pose~\cite{wang2024motionctrl,he2024cameractrl}, is also crucial.
GCD~\cite{vanhoorick2024gcd} firstly grants the video generation model to synthesize novel view video according to the assigned camera trajectory so that the director can recreate the footage he wants for a given video.
However, GCD needs to be trained on video data rendered from a simulator due to its training paradigm, which highly limits its generalization performance.
We bring the 4D Gaussian splatting~\cite{wu20244d} to a video diffusion transformer so that we can migrate the controllability of 4D Gaussian splatting, such as camera intrinsic and extrinsic and object editing, to video generation.


%% file: sec/3_method.tex
\section{Method}
\label{sec:method}

Achieving precise control over 4D content in video generation is essential for integrating this technology into film production.
To better accommodate video generation in cinematic studios, we tackle the 4D video control problem.
Inspired by classical Monocular Dynamic Video Synthesis (MDVS) methods~\cite{gao2022monocular,gao2021dynamic,wang2024shape} that render novel view videos by obtaining a 4D representation,
we propose a novel framework that learns 4D video control with the help of pseudo 4D Gaussian fields.
In contrast to GCD~\cite{vanhoorick2024gcd} which learns multi-camera shooting video generation from multi-view synchronized videos, our proposed framework directly learns from conventional monocular videos and can be naturally scaled up along with the base video Diffusion Transformers (DiT).
In this section, we first introduce the efficient dense 3D point tracking (D3D-PT), which is the cornerstone of the pseudo-4D Gaussian fields in training, and then demonstrate the 4D Gaussian field construction.
Finally, we elaborate on how to fine-tune a pre-trained DiT to generate videos with 4D control.


%


%
\subsection{Dense 3D Point Tracking}
%

Given the video frames $\mathbf{I} \in \mathbb{R}^{T \times H \times W \times C}$ and the corresponding metric depth maps $\mathbf{D} \in \mathbb{R}^{T \times H \times W}$, the goal of dense 3D point tracking (D3D-PT) is to estimate the 3D trajectories and visibilities $\mathbf{v} \in \mathbb{R}^{T \times H \times W}$ of all query pixels on the first frame $\mathbf{x}_0 \in \mathbb{R}^{H \times W \times 2}$ throughout the video. 
The 3D point trajectories is parameterized as their 2D locations and depth on the images $\mathbf{x} \in \mathbb{R}^{T \times H \times W \times 2}, \mathbf{d} \in \mathbb{R}^{T \times H \times W}$.
Taking both spatial and temporal video information is necessary but directly learning dense 3D point tracking by encoding the whole video is too expensive.
Inspired by DOT~\cite{lemoing2024dense}, we propose to estimate dense 3D point tracking in two stages: initialize the 3D point tracks in the first stage by encoding sparse but longer temporal information and then iteratively refine them in the second stage by encoding dense pair-wise information.

\noindent \textbf{Initialization}. Following DOT~\cite{lemoing2024dense}, we estimate sparse 2D point-tracking with CoTracker~\cite{karaev23cotracker} and obtain the dense point tracking $\mathbf{x}_{t} \in \{\mathbf{x}_{0}, \dots, \mathbf{x}_{T-1}\}$ through interpolation.
With such 2D point tracking initialization, we further initialize the point depths in all frames $\{\mathbf{d}_{0}, \dots, \mathbf{d}_{T-1}\}$.
Intuitively, the point depth in the first frame is assigned according to its depth map.
As the point depth variations are derived from the first frame, we also initialize the point depth in the other frames by the depth of the first frame $\mathbf d_t=\mathbf d_0=\mathbf{D}_0(\mathbf{x}_0)$.
All of the initial point visibilities $\mathbf{v}_{t} \in \{\mathbf{v}_{0}, \dots, \mathbf{v}_{T-1}\}$ are set to $1$, indicating that the points are visible.
The initialized variables $\{\mathbf{x}_{0}, \dots, \mathbf{x}_{T-1}\}, \{\mathbf{d}_{0}, \dots, \mathbf{d}_{T-1}\}, \{\mathbf{v}_{0}, \dots, \mathbf{v}_{T-1}\}$ will be iteratively refined in the next stage.

\noindent \textbf{Dense 3D Point Tracking Refinement}. 
We sequentially refine the coarse initialization results for each target video frame.
Given the reference frame $\mathbf{I}_{0}$ and a target frame $\mathbf{I}_t$, we refine the point tracks $\mathbf x_t$ and visibilities 
$\mathbf v_t$ with a RAFT~\cite{teed2020raft} network.
Specifically, after computing the correlation pyramid $\mathbf{}$ from the image features of $\mathbf{I}_{0}$ and $\mathbf{I}_t$, we iterativelly predict the residual point tracks $\Delta \mathbf{x}_{t,i}$ and visibilities  $\Delta \mathbf{v}_{t,i}$ at the $i$-th iteration according to the cropped correlation information and update them $\mathbf x_{t,i}=\mathbf x_{t,i-1} + \Delta \mathbf{x}_{t,i}, \mathbf v_{t,i}=\mathbf v_{t,i-1} + \Delta \mathbf{v}_{t,i}$. 
Please refer to the supplementary for more details.
But how do we refine the point depth? Adopting RAFT-3D~\cite{teed2021raft} is not feasible because we do not accept the rigid body assumptions in conventional dynamic videos.
Another straightforward solution is to expand the RGB images to RGB-D images by concatenating the depth map to the RGB images and refining the point depth concurrently.
However, we rely on monocular depth prediction methods in the inference stage.
The distribution of predicted depth may mismatch the depth distribution used during training and such a tight-coupled design presents poor generalization performance, which is critical in the following video generation training.
To improve the generalization performance of 3D point tracking, we propose a loosely coupled point depth refinement module.
Given the depth map of the target frame $\mathbf D_t$, we build a 4-layer depth pyramid $\hat{\mathbf D}_t$ by downsampling the depth map 4 times.
In addition to the depth information of the current frame, we iteratively refine the point depth with the 2D point tracking features.
At the $i$-th iteration, given the point tracking $\mathbf x_{t,i}$ and its feature $\mathbf m_{t,i}, \mathbf{inp}, \mathbf{net}$ (see supplementray), we crop depth information $\hat{\mathbf{D}}_t(\mathbf x_{t,i})\in \mathbb{R}^{H\times W\times 9\times 9\times 4}$ centered at $\mathbf x_{t,i}$, encode them through a motion encoder $MotionEnc$:
\begin{equation}
    \mathbf{m}^{d}_{t,i} = MotionEnc(\mathbf{m}_{t,i}, {\mathbf{d}_{t,i-1}}, \hat{\mathbf{D}}_t(\mathbf{x}_{t,i})),
\end{equation}
and refine the point depth $\mathbf d_{t,i} = \mathbf d_{t,i-1} + \Delta \mathbf d_{t,i}$ with a recurrent network $ConvGRU$ and a lightweight convolutional network $DepthHead$:
\begin{equation}
\begin{aligned}
        \mathbf{net} & = ConvGRU(\mathbf{net}, \mathbf{inp}, \mathbf{m}^{d}_{t,i}), \\
        \Delta \mathbf d_{t,i} & = DepthHead( \mathbf{net}).
\end{aligned}
\end{equation}

In this way, our network refines the point depth prediction by reading depth information from $\hat{\mathbf{D}}_t(\mathbf x_{t,i})$ and generalizes well when the input depth distribution differs from the training depth.
Our 3D point tracking accuracy surpasses SpatialTracker and accelerates the speed by two orders of magnitude, which builds the pseudo 4D Gaussian fields for our training videos.

\subsection{Pseudo 4D Gaussian Field}

Inspired by typical Monocular Dynamic View Synthesis (MDVS) that renders novel view images via a 4D Gaussian field, we tackle the 4D video control problem with the help of 4D Gaussian fields.
Optimizing a 4D Gaussian field with geometric constraints is time-consuming, around 1 hour per video, and such a method is always limited in object-centric scenarios due to the canonical space assumption.
Our goal is to control the camera intrinsic, extrinsic, and object motion in open-world videos.
The 4D video control can support various lens languages, such as multi-camera shooting and dolly zoom.
Therefore, we need to build the Gaussian field efficiently and flexibly.
We observe that the 4D motion field can be simply represented by the movements of 3D points along the time, which motivates us to construct a pseudo 4D Gaussian field by the proposed efficient dense 3D point tracking.



Specifically, 
a 4D Gaussian field consists of $N=H\times W$ points have the following parameters: $\mathbf{s}\in \mathbb{R}^{N\times3}, \mathbf{\alpha} \in \mathbb{R}^{N\times3}, \mathbf p \in \mathbb{R}^{T\times N\times 3}, \mathbf c \in \mathbb{R}^{N\times 3}$ describing the scale, opacity, position, and color of Gaussians.
We create a Gaussian for each pixel in the source frame.
The scale and opacity are set as constants.
We estimate the dense 3D trajectories starting from the source video frame throughout all video frames and assign them to Gaussian positions $\mathbf{p}_t \in \mathbb{R}^{N \times 3} = (\mathbf{x}_t, \mathbf{d}_t)$ and a 3D Gaussian field at timestep $t$ can be directly extracted. 
The color of Gaussians is naturally assigned the RGB value of the corresponding pixel. 
Given a Gaussian field at timestep $t$ and transformed to the camera coordinate through a camera pose, we render the image $\hat{\mathbf I}$ following the standard Gaussian splatting~\cite{3d_gs}:

\begin{equation}
\hat{\mathbf I}=\sum_{i=0}^{N-1} c_i \alpha_i \prod_{j=1}^{i-1}\left(1-\alpha_j\right).
\end{equation}

We call this 4D Gaussian field ``pseudo'' because we do not optimize the parameters and thus the rendered images and videos inevitably contain artifacts.
Fortunately, we do not directly regard such rendered videos as our output but exploit them as guidance for video generation.
We provide an overview of GS-DiT in Fig.~\ref{fig:pseudo_4dgs}.
During training, the videos are rendered according to the same camera trajectories as the original training video, and our proposed GS-DiT learns to generate the original video from the rendered video.
The rendered video used as conditional signals provides temporal information with multi-view consistency but presents serious artifacts.
GS-DiT is encouraged to restore the artifacts by connecting the temporal guidance information and strong video prior.

In the inference stage, we can also construct a pseudo 4D Gaussian field and render the videos with novel camera trajectories to guide the video generation.
Besides dense 3D point tracking, the pseudo 4D Gaussian field can be constructed by other methods, \eg{}, directly lifting per-frame depth to the 3D space or optimizing a real 4D Gaussian field in a canonical space.
In this paper, we use the lift depth map as a major example because it is the most efficient way.
With the pseudo 4D Gaussian field design, our GS-DiT can generate various visual effects.
For example, based on the same Gaussian field, rendering videos with different camera trajectories generates multi-camera shooting videos, and simultaneously adjusting the camera intrinsic and extrinsic produces dolly zoom effects.





\begin{figure}[tb]
    \centering
    \includegraphics[trim={1.5cm 2.5cm 1.5cm 2cm},clip,width=0.49\textwidth]{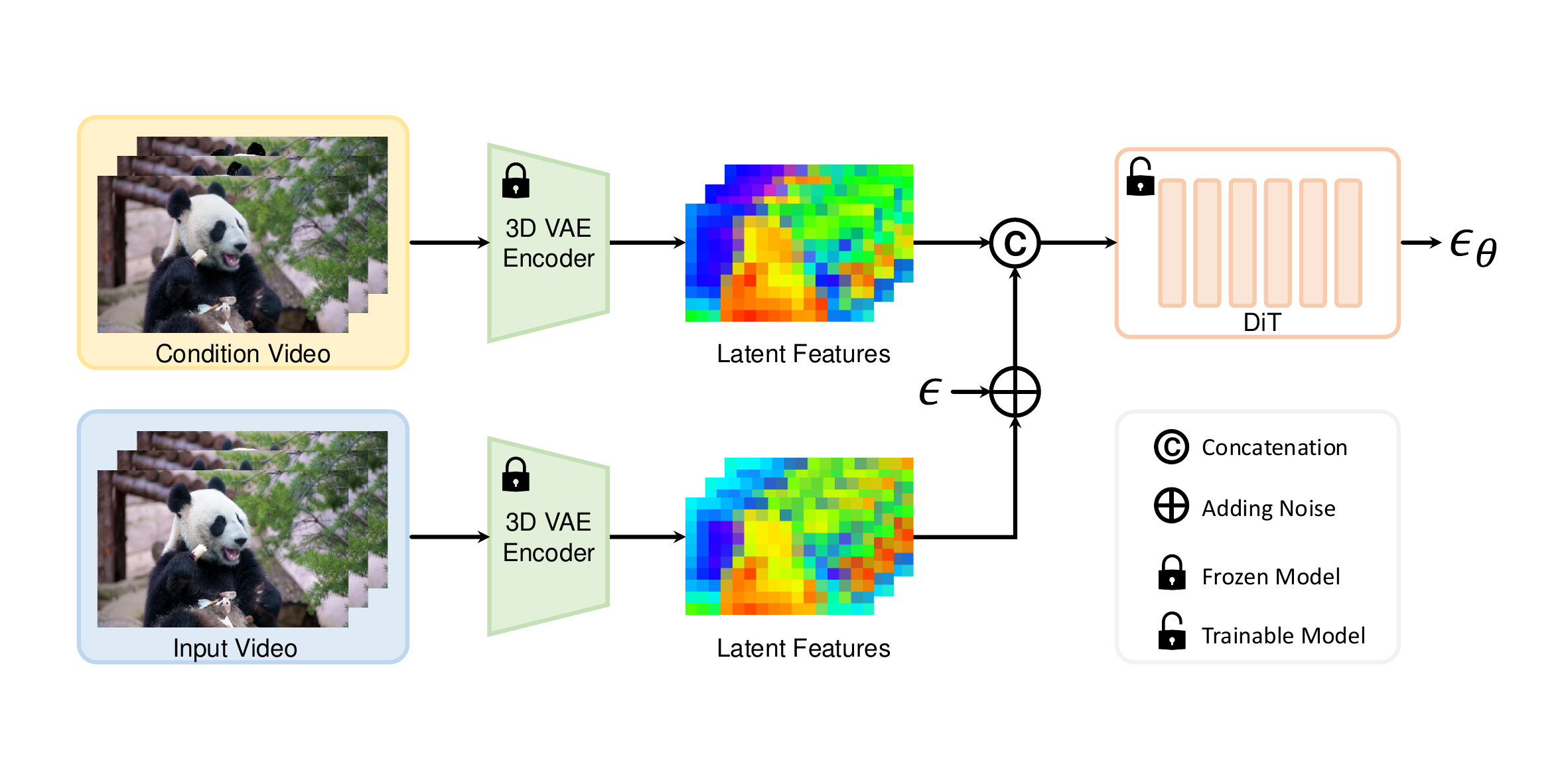}
    \vspace{-0.6cm}
    \caption{\textbf{The neural network architecture of GS-DiT.} GS-DiT generates video conditioned on the video rendered from our pseudo 4D Gaussian field.}
    \vspace{-0.5cm}
    \label{fig:dit_training}
\end{figure}

\begin{table*}[tb]
\scriptsize
\centering
\begin{tabular}{lcccccccccccc}
\toprule
\multirow{2}{*}{Methods} & \multicolumn{3}{c}{Kinetics} & \multicolumn{3}{c}{DAVIS} & \multicolumn{3}{c}{RGB-Stacking} & \multicolumn{3}{c}{Average} \\
& AJ $\uparrow$ & $< {\delta}^{x}_{avg} \uparrow$ & OA $\uparrow$ & AJ $\uparrow$ & $< {\delta}^{x}_{avg} \uparrow$ & OA $\uparrow$ & AJ $\uparrow$ & $< {\delta}^{x}_{avg} \uparrow$ & OA $\uparrow$ & AJ $\uparrow$ & $< {\delta}^{x}_{avg} \uparrow$ & OA $\uparrow$ \\
\midrule
TAP-Net~\cite{doersch2022tap} & 38.5 & 54.4 & 80.6 & 33.0 & 48.6 & 78.8 & 54.6 & 68.3 & 87.7 & 42.0 & 57.1 & 82.4 \\
PIPs~\cite{harley2022particle} & 31.7 & 53.7 & 72.9 & 42.2 & 64.8 & 77.7 & 15.7 & 28.4 & 77.1 & 29.9 & 50.0 & 75.9 \\
OmniMotion~\cite{wang2023omnimotion} & - & - & - & 46.4 & 62.7 & 85.3 & 69.5 & 82.5 & 90.3 & - & - & - \\
TAPIR~\cite{doersch2023tapir} & 49.6 & 64.2 & 85.0 & 56.2 & 70.0 & 86.5 & 54.2 & 69.8 & 84.4 & 53.3 & 68.0 & 85.3 \\
BootsTAPIR~\cite{doersch2024bootstap} & \textbf{54.6} & \textbf{68.4} & \textbf{86.5} & \textbf{61.4} & 73.6 & 88.7 & - & - & - & - & - & - \\
CoTracker~\cite{karaev23cotracker} & 48.7 & 64.3 & \textbf{86.5} & 60.6 & \textbf{75.4} & \textbf{89.3} & 63.1 & 77.0 & 87.8 & 57.4 & 72.2 & 87.8 \\
DOT~\cite{lemoing2024dense} & 48.4 & 63.8 & 85.2 & 60.1 & 74.5 & 89.0 & \textbf{77.1} & \textbf{87.7} & \textbf{93.3} & \textbf{61.9} & \textbf{75.3} & \textbf{89.2} \\
\midrule
SpatialTracker~\cite{SpatialTracker} & 50.1 & \textbf{65.9} & $\textbf{86.9}$ & 61.1 & $\textbf{76.3}$ & $\textbf{89.5}$ & 63.5 & 77.6 & 88.2 & 58.2 & 73.3 & 88.2 \\

Ours + ZoeDepth & \textbf{51.9} & 64.6 & 86.1 & \textbf{61.3} & 74.5 & 89.4 & \textbf{77.0} & \textbf{86.4} & \textbf{92.8} & $\textbf{63.4}$ & $\textbf{75.2}$ & $\textbf{89.4}$ \\
\bottomrule
\end{tabular}
\vspace{-0.3cm}
\caption{Comparison of 2D point tracking on TAPVid.}
\label{tab:sparse_2d_pt}
    \vspace{-0.2cm}
\end{table*}

\begin{table*}[tb]
\scriptsize
\centering
\begin{tabular}{lcccccccccccc}
\toprule
\multirow{2}{*}{Methods} & \multicolumn{3}{c}{Aria} & \multicolumn{3}{c}{DriveTrack} & \multicolumn{3}{c}{PStudio} & \multicolumn{3}{c}{Average} \\
& 3D-AJ $\uparrow$ & APD $\uparrow$ & OA $\uparrow$ & 3D-AJ $\uparrow$ & APD $\uparrow$ & $\mathrm{OA} \uparrow$ & 3D-AJ $\uparrow$ & APD $\uparrow$ & $\mathrm{OA} \uparrow$ & 3D-AJ $\uparrow$ & APD $\uparrow$ & $\mathrm{OA} \uparrow$ \\
\midrule
BootsTAPIR~\cite{doersch2024bootstap} + ZoeDepth~\cite{bhat2023zoedepth} & 8.6 & 14.5 & 86.9 & 5.1 & 8.7 & \textbf{83.5} & \textbf{10.2} & \textbf{17.7} & \textbf{82.0} & 8.0 & 13.6 & \textbf{84.1} \\
SpatialTracker~\cite{SpatialTracker} & 9.2 & 15.1 & 89.9 & 5.8 & 10.2 & 82.0 & 9.8 & \textbf{17.7} & 78.4 & 8.3 & 14.3 & 83.4 \\
Ours + ZoeDepth\cite{bhat2023zoedepth} & \textbf{10.0} & \textbf{16.1} & \textbf{90.0} & \textbf{7.2} & \textbf{12.0} & 81.1 & 9.8 & 17.3 & 80.5 & \textbf{9.0} & \textbf{15.1} & 83.9 \\
\midrule
Ours + Depth Anything V2~\cite{depth_anything_v2} & 14.5 & 21.9 & \textbf{90.0} & 8.6 & 13.8 & 81.1 & \textbf{11.3} & \textbf{19.4} & \textbf{80.5} & \textbf{11.4} & \textbf{18.3} & \textbf{83.9} \\
Ours + UniDepth V2~\cite{piccinelli2024unidepth} & \textbf{15.0} & \textbf{22.2} & \textbf{90.0} & \textbf{11.6} & \textbf{18.1} & \textbf{81.1} & 6.6 & 12.2 & \textbf{80.5} & 11.0 & 17.5 & \textbf{83.9} \\
\bottomrule
\end{tabular}
\vspace{-0.3cm}
\caption{Comparison of 3D point tracking on TAPVid-3D minival split.}

\label{tab:sparse_3d_pt}
\end{table*}

\begin{table*}[tb]
\scriptsize
\centering
\begin{tabular}{lccccccccc}
\toprule
\multirow{2}{*}{Methods} & \multicolumn{3}{c}{DAVIS} & \multicolumn{3}{c}{Sora} & \multicolumn{3}{c}{Pixabay}\\
 & PSNR $\uparrow$ & SSIM $\uparrow$ & LPIPS $\downarrow$ & PSNR $\uparrow$ & SSIM $\uparrow$ & LPIPS $\downarrow$  & PSNR $\uparrow$ & SSIM $\uparrow$ & LPIPS $\downarrow$ \\
\midrule
MonST3R & 14.12 & 0.59 & 0.31 & 15.32 & 0.59 & 0.30 & 19.78 & \textbf{0.74} & 0.22 \\
GCD & 15.04 & 0.41 & 0.48 & 11.96 & 0.32 & 0.52 & 13.71 & 0.42 & 0.50 \\
Ours  & \textbf{19.18} & \textbf{0.60} & \textbf{0.23} & \textbf{17.92} & \textbf{0.60} & \textbf{0.20} & \textbf{22.66} & 0.73 & \textbf{0.15} \\
\bottomrule
\end{tabular}
\vspace{-0.3cm}
\caption{Comparison of the multi-shooting video generation quality.}
\label{tab:recon_comparison}
\end{table*}

\subsection{Video Generation with GS-DiT}

We obtain our GS-DiT that controls 4D content in videos by finetuning the pre-trained 4D CogVideoX~\cite{yang2024cogvideox}, an open-source DiT architecture video generation model. 
We first brief CogVideoX and then elaborate on how to finetune the pretrained CogVideoX as our GS-DiT.

\noindent \textbf{CogVideoX}. 
CogVideoX contains three parts: a T5 text encoder~\cite{raffel2020exploring}, a 3D causal VAE that compresses 4 RGB video frames into 1 latent feature frame of 16 channels, and a full attention transformer backbone to encode all information.
CogVideoX generates a video consisting of 49 frames from text.
The training video is first compressed into 13 latent feature frames by the 3D causal VAE, where three dummy frames are padded in front, and diffusion training is conducted in the latent space with a full attention transformer.


\noindent \textbf{GS-DiT} is designed to generate a video from a guidance video. The guidance video is rendered from the pseudo 4D Gaussian field with a given camera trajectory, so the camera trajectory information is implicitly represented by the rendered video.
In this way, GS-DiT can generate multi-camera shooting videos.
Specifically, we render 13 video frames as the guidance.
To inject the guidance video information into DiT, we pad 3 dummy frames in front of each frame and individually encode the 4-frame group into a 16-channel latent feature frame with the 3D VAE.
Such conditional latent features are concatenated to the corresponding noise features, which constitute 32-channel features.
The patch embedding layer used in the attention is expanded to accept the 32-channel features and we initialize the additional convolution parameters by 0.

%
Following CogVideoX, we finetune our model $\theta$ with the standard DDPM~\cite{ho2020denoising} formulation:
\begin{equation}
L(\theta):=\mathbf{E}_{t, \mathbf{I}, \epsilon}\left\|\epsilon-\epsilon_\theta\left(\sqrt{\bar{\alpha}_t} \mathbf{I}+\sqrt{1-\bar{\alpha}_t} \epsilon, \hat{\mathbf{I}}, t\right)\right\|^2.
\end{equation}
We slightly abuse the annotation $t$ and $\alpha$, which indicates the denoising timestamp and coefficients in the standard DDPM here. $\epsilon \in \mathcal{N}(0,1)$ is the standard Gaussian noise and $\epsilon_{\theta}$ is the prediction of our GS-DiT.

%% file: sec/4_exp.tex
\section{Experiments}
\label{sec:exp}


\begin{figure*}[tb]
    \centering
    \includegraphics[trim={0cm, 1.5cm, 0cm, 1cm}, clip, width=1.0\linewidth]{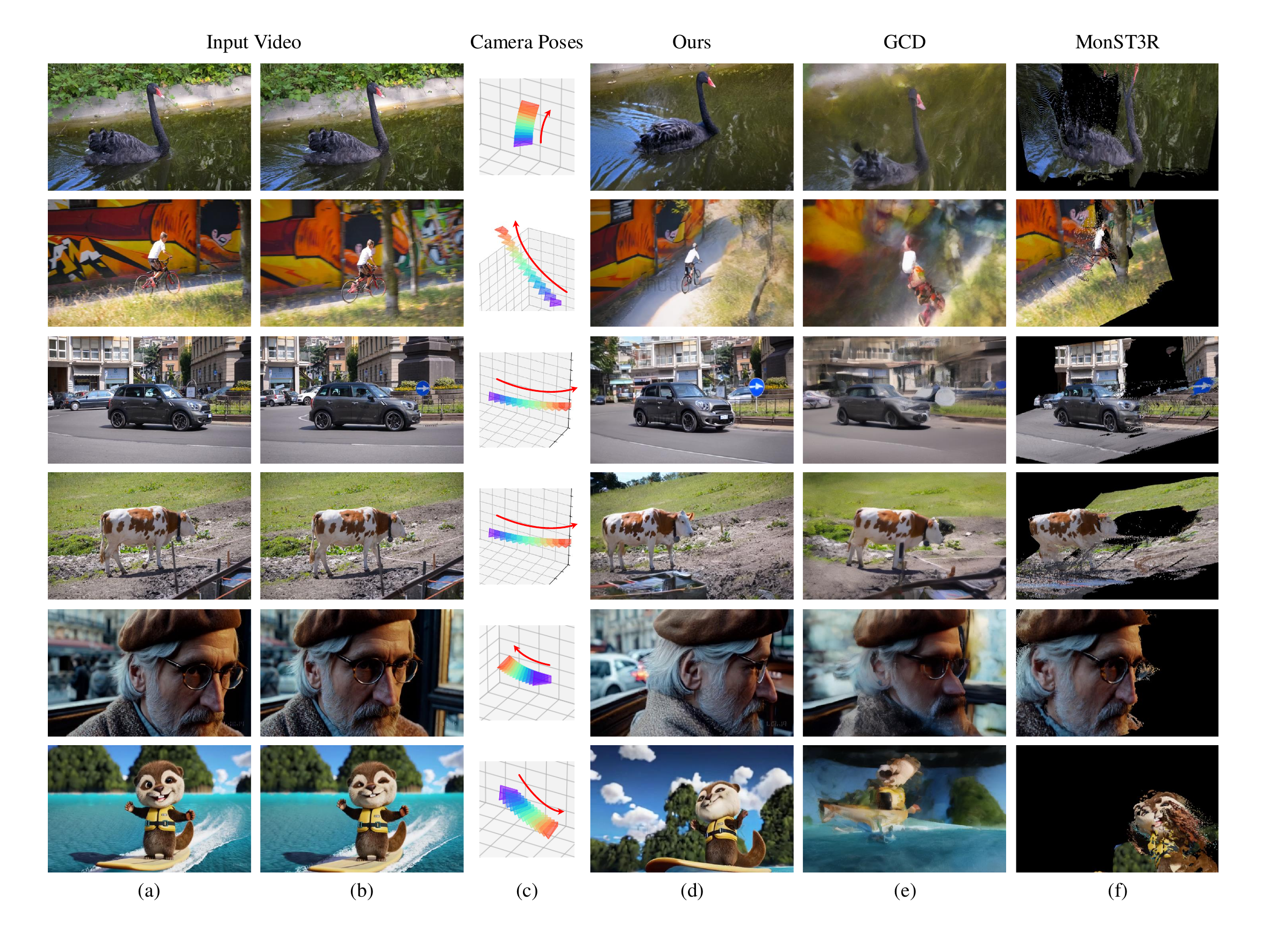}
    \vspace{-0.6cm}
    \caption{\textbf{Qualitative comparison of synchronized video generation with camera control}. (a)(b) are the first and last frame of the input video. (c) is the camera trajectory visualization. (d)(e)(f) are the last frame of the video generated by our GS-DiT, GCD, and MonST3R.}
    \label{fig:enter-label}
        \vspace{-0.5cm}
\end{figure*}

We first evaluate our proposed dense 3D point tracking (D3D-PT) and then evaluate our 4D video control framework GS-DiT.
Finally, we show the controllability of GS-DiT with dolly zoom and object motion editing.
\textit{We strongly suggest readers refer to the supplemented video}.

\subsection{Dense 3D Point Tracking}
As there are no available dense 3D point tracking benchmarks, we evaluate our dense 3D point tracking method on the sparse 3D point tracking benchmark TAPVid-3D~\cite{koppula2024tapvid} and sparse 2D point tracking benchmark TAPVid~\cite{doersch2022tap}. To our best knowledge, we are the first that address the dense 3D point tracking problem. D3D-PT achieves state-of-the-art performance even compared with the sparse 3D point tracking methods and sparse 2D point tracking methods. Nonetheless, D3D-PT presents extreme efficiency that accelerates 3D point tracking of dense pixels by $\sim 90$ times.

\noindent \textbf{Implementation Details}
We train our D3D-PT model on 8 NVIDIA A100-SXM4-80GB GPUs for 500,000 iterations. Our training data is generated with the Kubric~\cite{greff2021kubric} simulator. 
We generate 11,000 24-frame RGB-D sequences in total at $512 \times 512$ resolution with corresponding dense 3D point tracking ground truth. We set the batch size to $64$ and the learning rate to $1 \times 10^{-4}$. 


\noindent \textbf{Experimental Setup}
We follow TAPVid and TAPVid-3D  to evaluate sparse 2D and 3D point tracking.
TAPVid consists of three sub-datasets Kinetics, DAVIS, and RGB-Stacking. Kinetics contains 1144 250-frame videos, covering various human-related actions. DAVIS contains 30 real videos ranging in length from 34 to 104. RGB-Stacking contains 50 synthetic 250-frame robotic stacking videos. 
%
We adopt the average Jaccard (AJ), $< {\delta}^{x}_{avg}$, and the occlusion accuracy (OA) evaluation metrics for sparse 2D point tracking. $< {\delta}^{x}_{avg}$ evaluates the average fraction of visible points that are within pixel thresholds ${\delta} \in \{1, 2, 4, 8, 16\}$ of their ground truth. 
%
TAPVid-3D includes three tracks: Aria, DriveTrack, and PStudio. We use the ``minival" split, where each track has 50 videos. Aria contains 300-frame real videos recorded in indoor scenes. DriveTrack contains outdoor driving videos with lengths from 25 to 300. PStudio contains 150-frame indoor real videos focusing on human motions. We also adopt 3D metrics to measure the quality of the predicted 3D point trajectories (APD), point visibility (OA), and both simultaneously (3D-AJ~\cite{koppula2024tapvid}). 
%

\begin{figure*}[tb]
    \centering
    \includegraphics[width=1.0\linewidth]{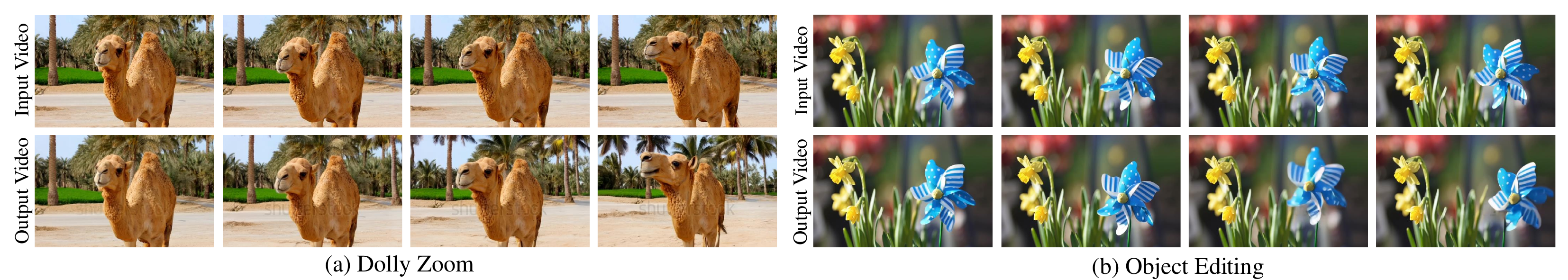}
    \vspace{-0.6cm}
    \caption{\textbf{Video generation with 4D control}. (a) presents the dolly zoom effects and (b) rotates the body of the fan.}
    \label{fig:obj_editing}
    \vspace{-0.5cm}
\end{figure*}

\noindent \textbf{Comparison on 2D Point Tracking}
We compare our D3D-PT with representative sparse 2D point tracking methods.
The most related works are DOT and SpatialTracker.
As shown in Tab.~\ref{tab:sparse_2d_pt}, even though our method is designed for dense 3D point tracking, it can also achieve state-of-the-art performance on 2D sparse point tracking benchmarks.
BootsTAPIR ranks 1st on Kinetics and DAVIS benchmarks because it is carefully improved with the bootstrap technique based on TAPIR.
Our method, which is not improved by bootstrap, ranks 2nd on the benchmarks and significantly outperforms BootsTAPIR's base model TAPIR.
Our method outperforms DOT, the dense 2D point tracking method, by 7.2\% on the AJ of Kinetics.
Our method also consistently outperforms the most recent sparse 3D point tracking method, SpatialTracker, in AJ on all of the benchmarks.

\noindent \textbf{Comparison on 3D Point Tracking} 
We compare our D3D-PT with BootsTAPIR + ZoeDepth and SpatialTracker in Tab.~\ref{tab:sparse_3d_pt}.
Notice that our method predicts dense 3D point tracking while only the sparse points that are annotated with ground truth are evaluated. 
The TAPVid-3D benchmark officially provides the results of BootsTAPIR + ZoeDepth, which combines the state-of-the-art sparse 2D point tracker BootsTAPIR and metric depth estimator ZoeDepth.
SpatialTracker is a pioneering method specifically designed for sparse 3D point tracking.
For a fair comparison, we use ZoeDepth as our depth estimator (Ours + ZoeDepth).
Our method significantly outperforms both methods on Aria and DriveTrack in terms of 3D-AJ and APD.

\noindent \textbf{Comparison on inference time}
We compare the inference time on the DAVIS dataset, which contains 90 480P videos with an average length of 69 (from 24 to 104).
We estimate dense 3D point tracking of the entire video on DAVIS starting from the first frame. 
We report the average time required to estimate a single frame.
The baseline method SpatialTracker cannot directly estimate dense 3D point tracking, so we split all query points into multiple chunks according to the default $50 \times 50$ sparse grid and estimate the total inference time to process all pixels.
SpatialTracker takes 89.8 seconds when processing a frame. Such an expensive time consumption prohibits it from processing large-scale video data.
Our method reduces the time overhead by two orders of magnitude.
Moreover, as shown in Tab.~\ref{tab:sparse_3d_pt}, our method also outperforms SpatialTracker in accuracy.
We, therefore, utilize our D3D-PT to construct the pseudo 4D Gaussian field in the following GS-DiT training.

\subsection{4D Video Control}

Multi-camera shooting video generation requires camera extrinsic control.
We quantitatively and qualitatively evaluate our GS-DiT in multi-camera shooting video generation.
We also show that we can control the camera intrinsic and object motion in videos with GS-DiT.

\noindent \textbf{Implementation Details}
We train our GS-DiT model on the WebVid-10M~\cite{Bain21} dataset. We randomly select 400K video clips with a length of 49 and stride 2 as the training data. 
We estimate the dense 3D point tracking with our D3D-PT and build a pseudo 4D Gaussian field for each video. Then we render the guidance videos from the pseudo 4D Gaussians to obtain the training set.
We train our GS-DiT for 100,000 steps on 8 NVIDIA A100-SXM4-80GB GPUs. We set batch size as 8, learning rate as $3 \times 10^{-5}$, and the resolution as $320 \times 512$.
%


\noindent \textbf{Experimental Setup}
Our GS-DiT bridges the 4D Gaussian splatting and the video generation.
We select GCD~\cite{vanhoorick2024gcd} and MonST3R~\cite{zhang2024monst3r} to compare the video quality and camera pose control.
GCD is the most relevant work that generates novel view videos while keeping the contents of the original video.
Its generalization performance is poor because the training data is generated by a simulator.
MonST3R is an efficient 4D video reconstruction method.
We build the 4D Gaussian field with MonST3R for the input video and render images through the test camera poses for evaluation.

\noindent \textbf{Quantitative Comparison on Multi-camera Shooting Video Generation}. We evaluate the video quality and camera controllability of video generation models on open-world dynamic videos. 
An ideal benchmark should be established by capturing the same dynamic scenes with multiple synchronized cameras but it is too expensive in the real open world.
Instead, given a monocular dynamic video, we design a special Arcball Rotation camera trajectory: starting from the identity pose, rotating the camera to $30^\circ$ in the middle frame, and then rotating back to the identity pose in the last frame. 
The rotation is orbit movement around the $(0,0,2)$ point.
For this kind of rotational camera motion, we set eight motion directions: left, right, up, down, upper left, lower left, upper right, and lower right.
Given a video and a set of camera trajectories for evaluation, the first and last frame of the generated video should be identical to the original video according to the camera control, but the content in the middle segment has corresponding viewpoint changes.
We evaluate the video generation controllability and quality by computing the visual alignment of the last frame between the original video and the generated video.
Therefore,
we build three datasets: DAVIS, Sora, and Pixabay, for evaluation.
DAVIS contains the same 30 real videos as those selected in the TAPViD benchmark.
Sora contains 15 generated videos collected from its technical report.
Pixabay contains 40 real videos collected from Pixabay's website.
We use the standard SSIM, PSNR, and LPIPS metrics for evaluation.
For each method, we generate 240, 120, and 320 videos respectively on DAVIS, Sora, and Pixabay for evaluation.
As shown in Tab.~\ref{tab:recon_comparison}, our method surpasses GCD and MonST3R all-sided.
Actually, GCD is almost collapsed due to its poor generalization performance.

\noindent \textbf{Qualitative Comparison on Camera Pose Control}
We qualitatively compare our method with GCD and MonST3R in Fig.~\ref{fig:enter-label}. There are six videos drawn from DAVIS and Sora datasets covering animals, humans, and vehicles. 
We design different camera trajectories as queries. 
For each method, we re-generate the input video according to the query camera trajectories as the visualization.
The videos generated by our GS-DiT accurately respond to the query camera trajectory, keep pleasing visual quality, and synchronize the object motion of the original video.
In contrast, GCD roughly aligns the camera trajectory and presents poor visual quality, which presents poor generalization performance. Due to the collapse of the visual quality, we can not tell whether the generated video is synchronized to the original video.
In the 2nd, 3rd, 4th, and 6th row, MonST3R presents good camera pose controllability but the visual quality is corrupted because some areas are unobserved.

\noindent \textbf{Camera Intrinsic and Object Motion Control}
Besides generating videos with camera extrinsic control, we can also control the camera intrinsic and edit the 4D Gaussian field when rendering the video.
As shown in Fig.~\ref{fig:obj_editing}, we apply the dolly zoom effects to the camel video (a) by simultaneously controlling the camera intrinsic and extrinsic. 
We also edit the pseudo Gaussian field derived from the video containing a rotating fan (b).
We apply a body rotation to the fan and generate the final video with GS-DiT.
The generated video obtains the original rotation movement of the input video and also presents the body rotation assigned by our control.
There are no floater artifacts after the object editing and the final video is natural.

%% file: sec/5-conclusion.tex
\section{Conclusion}
We have proposed a framework for 4D video control. By bringing pseudo Gaussian fields to video generation, we have endowed our GS-DiT with the 4D controllability from Gaussian fields, such as camera intrinsic and extrinsic editing and object motion editing.
Since our 4D representation is not optimized, we look forward to generating real 4D video with our GS-DiT in the future.

%% file: sec/supp.tex
\section{More Details about D3D-PT}

RAFT~\cite{teed2021raft} is designed for optical flow estimation, \ie{}, regressing a per-pixel displacement field $\mathbf{f}_t: \mathbb{I}^{H\times W\times 2} \rightarrow \mathbb{R}^{H\times W\times 2}$ that maps each source pixel to the coordinate $\mathbf x_t$ in the target video frame $t$.
Suppose the coordinate of the pixels in the source image is $\mathbf x_0$, the target coordinate can be derived from $\mathbf x_t = \mathbf x_0 + \mathbf f_t$.
DOT~\cite{lemoing2024dense} adopt RAFT to refine the dense point tracking.
Drawing inspiration from RAFT and DOT, we design the network architecture of our D3D-PT by iteratively refining the dense 3D point tracking, including the 2D point tracking $\mathbf x_t$, the visibility $\mathbf v_t$, and the depth $\mathbf d_t$.
We refine the 2D point tracking $\mathbf x_t$ and the visibility  $\mathbf v_t$ following RAFT, and refine the depth $\mathbf d_t$ with a recurrent decoder that is loosely coupled to the 2D point tracking decoder.


RAFT encodes image features with shallow CNNs, computes correlation volumes for all pairs of pixel features, builds 4-layer correlation pyramids by average pooling, and iteratively refines the correspondence estimation with a recurrent decoder according to the correlation pyramids and the image features.
Specifically, given a pair of RGB images, RAFT encodes them with a siamese network as $\mathbf{feat}_0\in \mathbb{R}^{H\times W\times C}$ and $\mathbf{feat}_t\in \mathbb{R}^{H\times W\times C}$ corresponding to the source image and the target image. $H, W, C$ denote the height, width, and channels of the encoded feature map. The source image is additionally encoded with a CNN to provide the context information $\mathbf{inp}\in \mathbb{R}^{H\times W\times C/2}$.
With the correlation pyramid $\mathbf{Corr}$ built from $\mathbf{feat}_0$ and $\mathbf{feat}_t$,
RAFT encodes the motion feature $\mathbf{m}_{t,i}$ from the flow and visibility estimated at the last iteration $\mathbf f_{t,i-1}, \mathbf v_{t,i-1}$, and cropped correlation information $\mathbf{Corr}(\mathbf x_{t,i-1})$:

\begin{equation}
\begin{aligned}
    & \mathbf{m}_{t,i}={MotionEnc_{flow}}(\mathbf f_{t,i-1}, \mathbf{Corr}(\mathbf x_{t,i-1}), \mathbf v_{t,i-1}).
\end{aligned}
\label{Eq: motion encoder}
\end{equation}

\begin{figure}[t]
    \centering
    \includegraphics[width=1.0\linewidth]{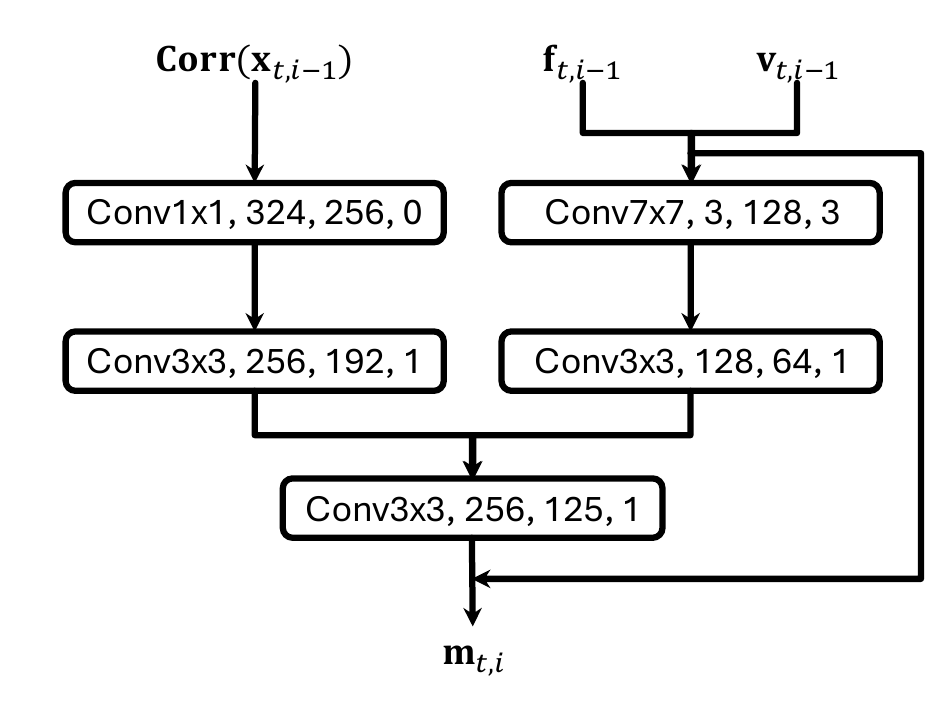}
    \caption{The structure of ${MotionEnc_{flow}}$.}
    \label{fig:motion_enc}
\end{figure}

We show the structure of the motion encoder in Fig.~\ref{fig:motion_enc}. There is a ReLU~\cite{agarap2018deep} activation function between the convolution layers.
The motion features $\mathbf{m}_{t,i}$ will be fed to the $ConvGRU_{flow}$ to estimate the flow update:

\begin{equation}
\begin{aligned}
    \mathbf{net}_{rgb,i} & = {ConvGRU_{flow}}(\mathbf{m}_{t,i}, \mathbf{net}_{rgb,i-1}, \mathbf{inp}), \\
    \Delta \mathbf f_{t,i} & = FlowHead( \mathbf{net}_{rgb,i}), \\
    \Delta \mathbf v_{t,i} & = VisHead( \mathbf{net}_{rgb,i}). 
\end{aligned}
\label{Eq: motion encoder}
\end{equation}

\begin{figure}[t]
    \centering
    \includegraphics[width=1.0\linewidth]{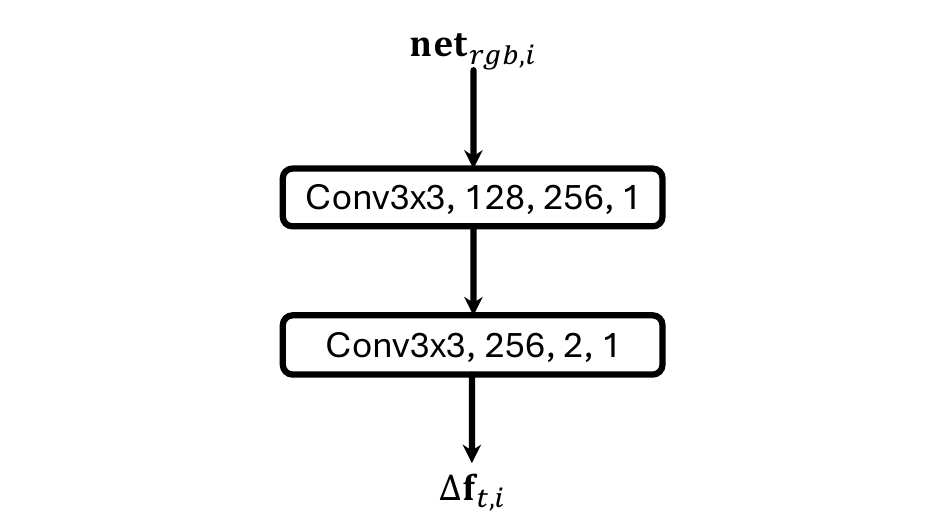}
    \caption{The structure of ${FlowHead}$.}
    \label{fig:flow_head}
\end{figure}

\begin{figure}[t]
    \centering
    \includegraphics[width=1.0\linewidth]{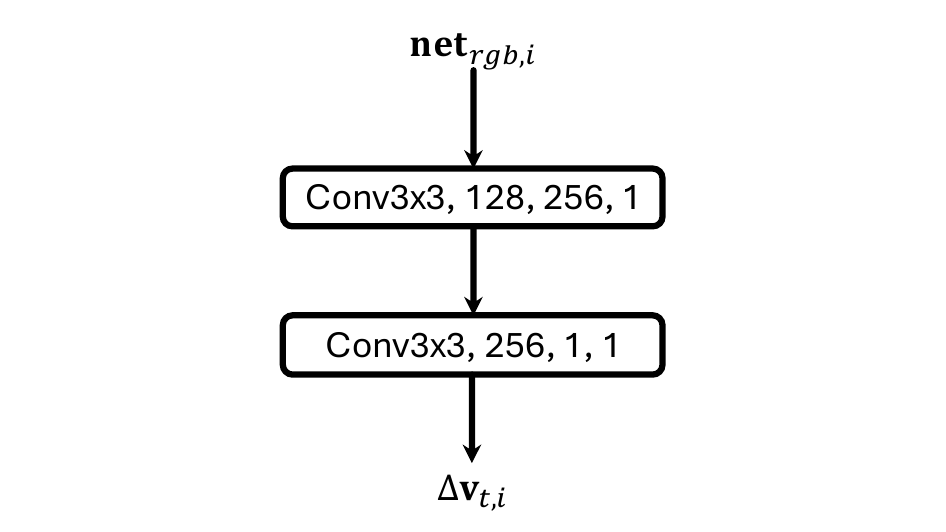}
    \caption{The structure of ${VisHead}$.}
    \label{fig:vis_head}
\end{figure}

\begin{figure}[t]
    \centering
    \includegraphics[width=1.0\linewidth]{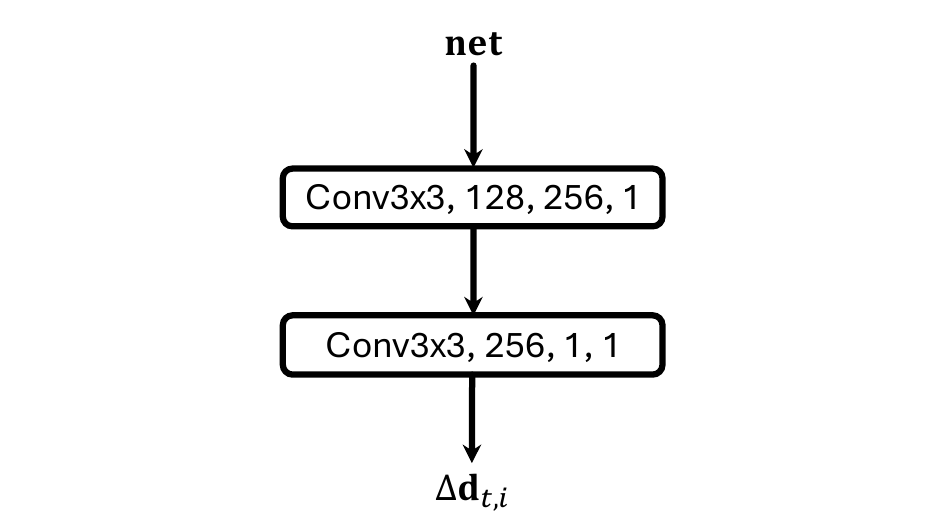}
    \caption{The structure of ${DepthHead}$.}
    \label{fig:depth_head}
\end{figure}

$MotionEnc_{flow}$, $ConvGRU_{flow}$, and $FlowHead$ are standard blocks used in DOT.
$\mathbf{net}_{rgb}$ is an iterativelly updated hidden state.
$ConvGRU_{flow}$ is the recurrent decoder used by RAFT. $FlowHead$, $VisHead$, and $DepthHead$ used to regress residual flow, visibility, and depth share similar structures.
We show them in Fig.~\ref{fig:flow_head}, Fig.~\ref{fig:vis_head}, and Fig.~\ref{fig:depth_head}.
Besides updating flow estimation, the motion feature $\mathbf{m}_{t,i}$ and the hidden feature $\mathbf{net}_{rgb,i-1}$ will also be used in the depth refinement as elaborated in the main paper.
We assign $\mathbf{net}_{rgb,i-1}$ to $\mathbf{net}$ used in the main paper.

\begin{figure*}[!t]
    \begin{center}
    \includegraphics[width=1.0\linewidth]{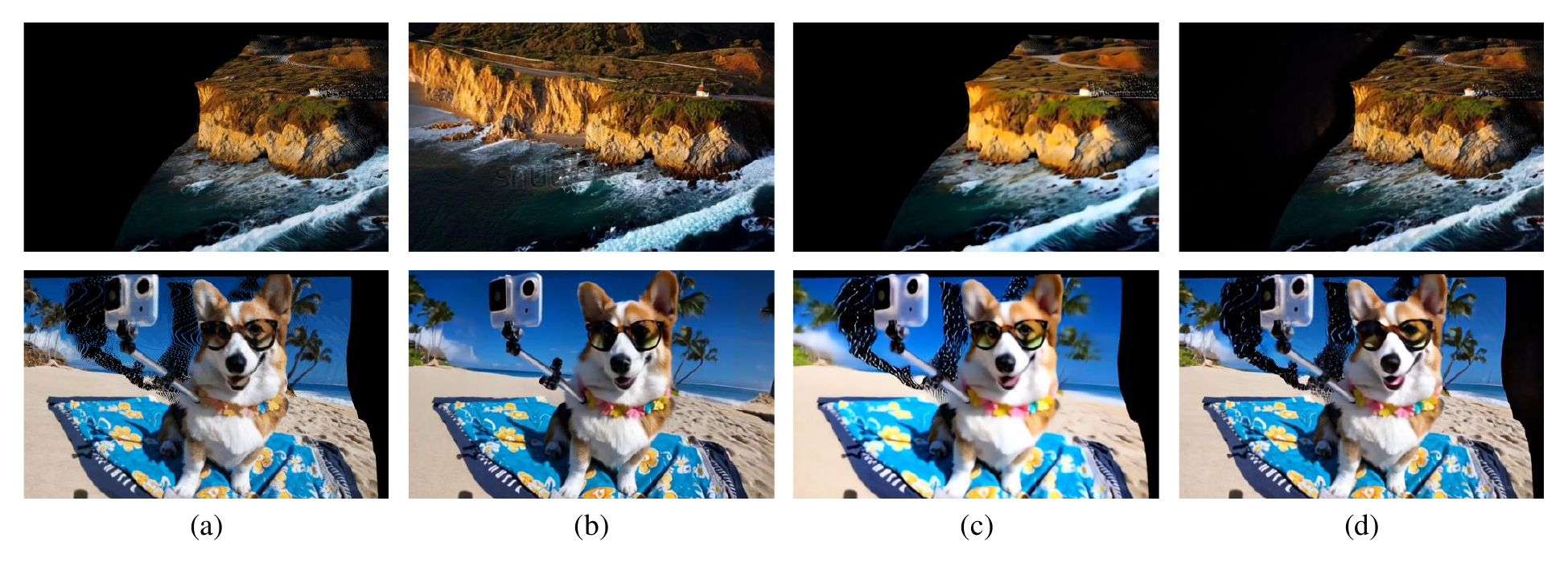}
    \end{center}
    \vspace{-0.6cm}
    \caption{Comparison with video inpainting. (a) is the input frame. (b) is the video frame generated by our GS-DiT. (c) and (d) are the video frames generated by Inpainting-A and Inpainting-B.
    }
    \label{fig:inpainting comparison}
\end{figure*}

\section{Ablation Study}

\begin{table}[tb]
\scriptsize
\centering
\begin{tabular}{lcccccccccccc}
\toprule
\multirow{2}{*}{Methods} &  \multicolumn{3}{c}{DAVIS} \\
& AJ $\uparrow$ & $< {\delta}^{x}_{avg} \uparrow$ & OA $\uparrow$  \\
\midrule
RGB-RAFT (DOT)~\cite{lemoing2024dense} & 60.1 & 74.5 & 89.0  \\
RGBD-RAFT  & 55.7 & 71.6 & 86.7 \\
D3D-PT (Ours)  & $\textbf{63.4}$ & $\textbf{75.2}$ & $\textbf{89.4}$ \\
\bottomrule
\end{tabular}
\caption{Ablation Study on DAVIS.}
\label{tab:sparse_2d_pt ablation}
\vspace{-0.6cm}
\end{table}

\noindent \textbf{Dense 3D Point Tracking}. We conduct an ablation study on DAVIS to show the superiority of our loosely coupled dense 3D point tracking design. As shown in Tab.~\ref{tab:sparse_2d_pt ablation}, DOT~\cite{lemoing2024dense} can be regarded as the baseline that takes the original RAFT to estimate the dense 2D point tracking.
Directly extending the RAFT to accept RGB-D images as inputs (RGBD-RAFT) degrades the point tracking accuracy seriously because the depth distribution in training is different from the distribution of depth estimated in the inference stage.
On the contrary, our loosely coupled 3D point tracking design D3D-PT improves the tracking accuracy on the DAVIS.

\noindent \textbf{GS-DiT}
GS-DiT generates video conditioned on the input video, which is rendered from the pseudo 4D Gaussian field.
The generated video is expected to fix the artifacts, such as the blurs and the incomplete areas, derived from the imperfect 4D Gaussians.
Such a process is similar to video inpainting, so we set a simple baseline that trains a DiT-based video inpainting model to reveal the essence of building the pseudo 4D Gaussian field for training.
We corrupted the videos with two simple masks: evenly distributed dispersed masks (Inpainting-A) and fixed-size grid masks at random locations (Inpainting-B).
We train all models with 5000 iterations at $320 \times 512$ resolution.
We expect that the masked region occupies 40\% of the images, so we set the dispersed occlusion ratio as 40\% in Inpainting-A and 
the grid mask with the size of $256 \times 256$ in Inpainting-B.
We remove part of the information to obtain the corrupted video according to the generated random mask as the condition video.
As shown in Fig.~\ref{fig:inpainting comparison}, our GS-DiT obtains clear details and infers reasonable unobserved regions.
In contrast, both inpainting models (Inpainting-A and Inpainting-B) fail to infer the incomplete regions and present blurry video frames. Moreover, the blurry effects is severe in the video frame generated by Inpainting-A.
This comparison shows that building the pseudo 4D Gaussian field with our D3D-PT is the cornerstone of the GS-DiT.

%% file: main.bbl
\begin{thebibliography}{57}
\providecommand{\natexlab}[1]{#1}
\providecommand{\url}[1]{\texttt{#1}}
\expandafter\ifx\csname urlstyle\endcsname\relax
  \providecommand{\doi}[1]{doi: #1}\else
  \providecommand{\doi}{doi: \begingroup \urlstyle{rm}\Url}\fi

\bibitem[Agarap(2018)]{agarap2018deep}
AF Agarap.
\newblock Deep learning using rectified linear units (relu).
\newblock \emph{arXiv preprint arXiv:1803.08375}, 2018.

\bibitem[Bain et~al.(2021)Bain, Nagrani, Varol, and Zisserman]{Bain21}
Max Bain, Arsha Nagrani, G{\"u}l Varol, and Andrew Zisserman.
\newblock Frozen in time: A joint video and image encoder for end-to-end retrieval.
\newblock In \emph{IEEE International Conference on Computer Vision}, 2021.

\bibitem[Bansal et~al.(2020)Bansal, Vo, Sheikh, Ramanan, and Narasimhan]{bansal20204d}
Aayush Bansal, Minh Vo, Yaser Sheikh, Deva Ramanan, and Srinivasa Narasimhan.
\newblock 4d visualization of dynamic events from unconstrained multi-view videos.
\newblock In \emph{Proceedings of the IEEE/CVF Conference on Computer Vision and Pattern Recognition}, pages 5366--5375, 2020.

\bibitem[Bhat et~al.(2023)Bhat, Birkl, Wofk, Wonka, and M{\"u}ller]{bhat2023zoedepth}
Shariq~Farooq Bhat, Reiner Birkl, Diana Wofk, Peter Wonka, and Matthias M{\"u}ller.
\newblock Zoedepth: Zero-shot transfer by combining relative and metric depth.
\newblock \emph{arXiv preprint arXiv:2302.12288}, 2023.

\bibitem[Bian et~al.(2023)Bian, Huang, Shi, Dong, Li, and Li]{bian2023contextpips}
Weikang Bian, Zhaoyang Huang, Xiaoyu Shi, Yitong Dong, Yijin Li, and Hongsheng Li.
\newblock Context-pips: Persistent independent particles demands spatial context features.
\newblock In \emph{Advances in Neural Information Processing Systems}, pages 55285--55298, 2023.

\bibitem[Blattmann et~al.(2023{\natexlab{a}})Blattmann, Dockhorn, Kulal, Mendelevitch, Kilian, Lorenz, Levi, English, Voleti, Letts, et~al.]{blattmann2023stable}
Andreas Blattmann, Tim Dockhorn, Sumith Kulal, Daniel Mendelevitch, Maciej Kilian, Dominik Lorenz, Yam Levi, Zion English, Vikram Voleti, Adam Letts, et~al.
\newblock Stable video diffusion: Scaling latent video diffusion models to large datasets.
\newblock \emph{arXiv preprint arXiv:2311.15127}, 2023{\natexlab{a}}.

\bibitem[Blattmann et~al.(2023{\natexlab{b}})Blattmann, Rombach, Ling, Dockhorn, Kim, Fidler, and Kreis]{blattmann2023align}
Andreas Blattmann, Robin Rombach, Huan Ling, Tim Dockhorn, Seung~Wook Kim, Sanja Fidler, and Karsten Kreis.
\newblock Align your latents: High-resolution video synthesis with latent diffusion models.
\newblock In \emph{Proceedings of the IEEE/CVF Conference on Computer Vision and Pattern Recognition}, pages 22563--22575, 2023{\natexlab{b}}.

\bibitem[Brooks et~al.(2024)Brooks, Peebles, Holmes, DePue, Guo, Jing, Schnurr, Taylor, Luhman, Luhman, Ng, Wang, and Ramesh]{videoworldsimulators2024}
Tim Brooks, Bill Peebles, Connor Holmes, Will DePue, Yufei Guo, Li Jing, David Schnurr, Joe Taylor, Troy Luhman, Eric Luhman, Clarence Ng, Ricky Wang, and Aditya Ramesh.
\newblock Video generation models as world simulators.
\newblock 2024.

\bibitem[Chen et~al.(2024)Chen, Wang, Wang, Wang, and Liu]{chen2024v3d}
Zilong Chen, Yikai Wang, Feng Wang, Zhengyi Wang, and Huaping Liu.
\newblock V3d: Video diffusion models are effective 3d generators.
\newblock \emph{arXiv preprint arXiv:2403.06738}, 2024.

\bibitem[Dhariwal and Nichol(2021)]{dhariwal2021diffusion}
Prafulla Dhariwal and Alexander Nichol.
\newblock Diffusion models beat gans on image synthesis.
\newblock \emph{Advances in neural information processing systems}, 34:\penalty0 8780--8794, 2021.

\bibitem[Doersch et~al.(2022)Doersch, Gupta, Markeeva, Recasens, Smaira, Aytar, Carreira, Zisserman, and Yang]{doersch2022tap}
Carl Doersch, Ankush Gupta, Larisa Markeeva, Adria Recasens, Lucas Smaira, Yusuf Aytar, Joao Carreira, Andrew Zisserman, and Yi Yang.
\newblock {TAP}-vid: A benchmark for tracking any point in a video.
\newblock \emph{Advances in Neural Information Processing Systems}, 35:\penalty0 13610--13626, 2022.

\bibitem[Doersch et~al.(2023)Doersch, Yang, Vecerik, Gokay, Gupta, Aytar, Carreira, and Zisserman]{doersch2023tapir}
Carl Doersch, Yi Yang, Mel Vecerik, Dilara Gokay, Ankush Gupta, Yusuf Aytar, Joao Carreira, and Andrew Zisserman.
\newblock Tapir: Tracking any point with per-frame initialization and temporal refinement.
\newblock \emph{ICCV}, 2023.

\bibitem[Doersch et~al.(2024)Doersch, Luc, Yang, Gokay, Koppula, Gupta, Heyward, Rocco, Goroshin, Carreira, and Zisserman]{doersch2024bootstap}
Carl Doersch, Pauline Luc, Yi Yang, Dilara Gokay, Skanda Koppula, Ankush Gupta, Joseph Heyward, Ignacio Rocco, Ross Goroshin, Jo{\~a}o Carreira, and Andrew Zisserman.
\newblock {BootsTAP}: Bootstrapped training for tracking-any-point.
\newblock \emph{Asian Conference on Computer Vision}, 2024.

\bibitem[Esser et~al.(2023)Esser, Chiu, Atighehchian, Granskog, and Germanidis]{esser2023structure}
Patrick Esser, Johnathan Chiu, Parmida Atighehchian, Jonathan Granskog, and Anastasis Germanidis.
\newblock Structure and content-guided video synthesis with diffusion models.
\newblock In \emph{Proceedings of the IEEE/CVF International Conference on Computer Vision}, pages 7346--7356, 2023.

\bibitem[Gao et~al.(2021)Gao, Saraf, Kopf, and Huang]{gao2021dynamic}
Chen Gao, Ayush Saraf, Johannes Kopf, and Jia-Bin Huang.
\newblock Dynamic view synthesis from dynamic monocular video.
\newblock In \emph{Proceedings of the IEEE/CVF International Conference on Computer Vision}, pages 5712--5721, 2021.

\bibitem[Gao et~al.(2022)Gao, Li, Tulsiani, Russell, and Kanazawa]{gao2022monocular}
Hang Gao, Ruilong Li, Shubham Tulsiani, Bryan Russell, and Angjoo Kanazawa.
\newblock Monocular dynamic view synthesis: A reality check.
\newblock \emph{Advances in Neural Information Processing Systems}, 35:\penalty0 33768--33780, 2022.

\bibitem[Garcia et~al.(2024)Garcia, Zeid, Schmidt, de~Geus, Hermans, and Leibe]{garcia2024fine}
Gonzalo~Martin Garcia, Karim~Abou Zeid, Christian Schmidt, Daan de Geus, Alexander Hermans, and Bastian Leibe.
\newblock Fine-tuning image-conditional diffusion models is easier than you think.
\newblock \emph{arXiv preprint arXiv:2409.11355}, 2024.

\bibitem[Greff et~al.(2022)Greff, Belletti, Beyer, Doersch, Du, Duckworth, Fleet, Gnanapragasam, Golemo, Herrmann, Kipf, Kundu, Lagun, Laradji, Liu, Meyer, Miao, Nowrouzezahrai, Oztireli, Pot, Radwan, Rebain, Sabour, Sajjadi, Sela, Sitzmann, Stone, Sun, Vora, Wang, Wu, Yi, Zhong, and Tagliasacchi]{greff2021kubric}
Klaus Greff, Francois Belletti, Lucas Beyer, Carl Doersch, Yilun Du, Daniel Duckworth, David~J Fleet, Dan Gnanapragasam, Florian Golemo, Charles Herrmann, Thomas Kipf, Abhijit Kundu, Dmitry Lagun, Issam Laradji, Hsueh-Ti~(Derek) Liu, Henning Meyer, Yishu Miao, Derek Nowrouzezahrai, Cengiz Oztireli, Etienne Pot, Noha Radwan, Daniel Rebain, Sara Sabour, Mehdi S.~M. Sajjadi, Matan Sela, Vincent Sitzmann, Austin Stone, Deqing Sun, Suhani Vora, Ziyu Wang, Tianhao Wu, Kwang~Moo Yi, Fangcheng Zhong, and Andrea Tagliasacchi.
\newblock Kubric: a scalable dataset generator.
\newblock 2022.

\bibitem[Guo et~al.(2024)Guo, Yang, Rao, Liang, Wang, Qiao, Agrawala, Lin, and Dai]{guo2023animatediff}
Yuwei Guo, Ceyuan Yang, Anyi Rao, Zhengyang Liang, Yaohui Wang, Yu Qiao, Maneesh Agrawala, Dahua Lin, and Bo Dai.
\newblock Animatediff: Animate your personalized text-to-image diffusion models without specific tuning.
\newblock \emph{International Conference on Learning Representations}, 2024.

\bibitem[Harley et~al.(2022)Harley, Fang, and Fragkiadaki]{harley2022particle}
Adam~W Harley, Zhaoyuan Fang, and Katerina Fragkiadaki.
\newblock Particle video revisited: Tracking through occlusions using point trajectories.
\newblock In \emph{ECCV}, 2022.

\bibitem[Harvey et~al.(2022)Harvey, Naderiparizi, Masrani, Weilbach, and Wood]{harvey2022flexible}
William Harvey, Saeid Naderiparizi, Vaden Masrani, Christian Weilbach, and Frank Wood.
\newblock Flexible diffusion modeling of long videos.
\newblock \emph{Advances in Neural Information Processing Systems}, 35:\penalty0 27953--27965, 2022.

\bibitem[He et~al.(2024)He, Xu, Guo, Wetzstein, Dai, Li, and Yang]{he2024cameractrl}
Hao He, Yinghao Xu, Yuwei Guo, Gordon Wetzstein, Bo Dai, Hongsheng Li, and Ceyuan Yang.
\newblock Cameractrl: Enabling camera control for text-to-video generation.
\newblock \emph{arXiv preprint arXiv:2404.02101}, 2024.

\bibitem[Ho and Salimans(2022)]{ho2022classifier}
Jonathan Ho and Tim Salimans.
\newblock Classifier-free diffusion guidance.
\newblock \emph{arXiv preprint arXiv:2207.12598}, 2022.

\bibitem[Ho et~al.(2020)Ho, Jain, and Abbeel]{ho2020denoising}
Jonathan Ho, Ajay Jain, and Pieter Abbeel.
\newblock Denoising diffusion probabilistic models.
\newblock \emph{Advances in neural information processing systems}, 33:\penalty0 6840--6851, 2020.

\bibitem[Ho et~al.(2022{\natexlab{a}})Ho, Chan, Saharia, Whang, Gao, Gritsenko, Kingma, Poole, Norouzi, Fleet, et~al.]{ho2022imagen}
Jonathan Ho, William Chan, Chitwan Saharia, Jay Whang, Ruiqi Gao, Alexey Gritsenko, Diederik~P Kingma, Ben Poole, Mohammad Norouzi, David~J Fleet, et~al.
\newblock Imagen video: High definition video generation with diffusion models.
\newblock \emph{arXiv preprint arXiv:2210.02303}, 2022{\natexlab{a}}.

\bibitem[Ho et~al.(2022{\natexlab{b}})Ho, Salimans, Gritsenko, Chan, Norouzi, and Fleet]{ho2022video}
Jonathan Ho, Tim Salimans, Alexey Gritsenko, William Chan, Mohammad Norouzi, and David~J Fleet.
\newblock Video diffusion models.
\newblock \emph{Advances in Neural Information Processing Systems}, 35:\penalty0 8633--8646, 2022{\natexlab{b}}.

\bibitem[Karaev et~al.(2024)Karaev, Rocco, Graham, Neverova, Vedaldi, and Rupprecht]{karaev23cotracker}
Nikita Karaev, Ignacio Rocco, Benjamin Graham, Natalia Neverova, Andrea Vedaldi, and Christian Rupprecht.
\newblock Cotracker: It is better to track together.
\newblock In \emph{Proc. {ECCV}}, 2024.

\bibitem[Karras et~al.(2022)Karras, Aittala, Aila, and Laine]{karras2022elucidating}
Tero Karras, Miika Aittala, Timo Aila, and Samuli Laine.
\newblock Elucidating the design space of diffusion-based generative models.
\newblock \emph{Advances in neural information processing systems}, 35:\penalty0 26565--26577, 2022.

\bibitem[Ke et~al.(2024)Ke, Obukhov, Huang, Metzger, Daudt, and Schindler]{ke2024repurposing}
Bingxin Ke, Anton Obukhov, Shengyu Huang, Nando Metzger, Rodrigo~Caye Daudt, and Konrad Schindler.
\newblock Repurposing diffusion-based image generators for monocular depth estimation.
\newblock In \emph{Proceedings of the IEEE/CVF Conference on Computer Vision and Pattern Recognition}, pages 9492--9502, 2024.

\bibitem[Kerbl et~al.(2023)Kerbl, Kopanas, Leimk{\"u}hler, and Drettakis]{3d_gs}
Bernhard Kerbl, Georgios Kopanas, Thomas Leimk{\"u}hler, and George Drettakis.
\newblock 3d gaussian splatting for real-time radiance field rendering.
\newblock \emph{ACM Trans. Graph.}, 42\penalty0 (4):\penalty0 139--1, 2023.

\bibitem[Koppula et~al.(2024)Koppula, Rocco, Yang, Heyward, Carreira, Zisserman, Brostow, and Doersch]{koppula2024tapvid}
Skanda Koppula, Ignacio Rocco, Yi Yang, Joe Heyward, Jo{\~a}o Carreira, Andrew Zisserman, Gabriel Brostow, and Carl Doersch.
\newblock {TAPVid}-{3D}: A benchmark for tracking any point in {3D}.
\newblock \emph{Advances in Neural Information Processing Systems}, 2024.

\bibitem[Le~Moing et~al.(2024)Le~Moing, Ponce, and Schmid]{lemoing2024dense}
Guillaume Le~Moing, Jean Ponce, and Cordelia Schmid.
\newblock Dense optical tracking: Connecting the dots.
\newblock In \emph{CVPR}, 2024.

\bibitem[Liu et~al.(2022)Liu, Gong, and Liu]{liu2022flow}
Xingchao Liu, Chengyue Gong, and Qiang Liu.
\newblock Flow straight and fast: Learning to generate and transfer data with rectified flow.
\newblock \emph{arXiv preprint arXiv:2209.03003}, 2022.

\bibitem[Nichol and Dhariwal(2021)]{nichol2021improved}
Alexander~Quinn Nichol and Prafulla Dhariwal.
\newblock Improved denoising diffusion probabilistic models.
\newblock In \emph{International conference on machine learning}, pages 8162--8171. PMLR, 2021.

\bibitem[Ozguroglu et~al.(2024)Ozguroglu, Liu, Sur{\'\i}s, Chen, Dave, Tokmakov, and Vondrick]{ozguroglu2024pix2gestalt}
Ege Ozguroglu, Ruoshi Liu, D{\'\i}dac Sur{\'\i}s, Dian Chen, Achal Dave, Pavel Tokmakov, and Carl Vondrick.
\newblock pix2gestalt: Amodal segmentation by synthesizing wholes.
\newblock In \emph{2024 IEEE/CVF Conference on Computer Vision and Pattern Recognition (CVPR)}, pages 3931--3940. IEEE Computer Society, 2024.

\bibitem[Peebles and Xie(2023)]{peebles2023scalable}
William Peebles and Saining Xie.
\newblock Scalable diffusion models with transformers.
\newblock In \emph{Proceedings of the IEEE/CVF International Conference on Computer Vision}, pages 4195--4205, 2023.

\bibitem[Piccinelli et~al.(2024)Piccinelli, Yang, Sakaridis, Segu, Li, Van~Gool, and Yu]{piccinelli2024unidepth}
Luigi Piccinelli, Yung-Hsu Yang, Christos Sakaridis, Mattia Segu, Siyuan Li, Luc Van~Gool, and Fisher Yu.
\newblock {U}ni{D}epth: Universal monocular metric depth estimation.
\newblock In \emph{Proceedings of the IEEE/CVF Conference on Computer Vision and Pattern Recognition (CVPR)}, 2024.

\bibitem[Poole et~al.(2022)Poole, Jain, Barron, and Mildenhall]{poole2022dreamfusion}
Ben Poole, Ajay Jain, Jonathan~T Barron, and Ben Mildenhall.
\newblock Dreamfusion: Text-to-3d using 2d diffusion.
\newblock \emph{arXiv preprint arXiv:2209.14988}, 2022.

\bibitem[Raffel et~al.(2020)Raffel, Shazeer, Roberts, Lee, Narang, Matena, Zhou, Li, and Liu]{raffel2020exploring}
Colin Raffel, Noam Shazeer, Adam Roberts, Katherine Lee, Sharan Narang, Michael Matena, Yanqi Zhou, Wei Li, and Peter~J Liu.
\newblock Exploring the limits of transfer learning with a unified text-to-text transformer.
\newblock \emph{Journal of machine learning research}, 21\penalty0 (140):\penalty0 1--67, 2020.

\bibitem[Rombach et~al.(2022)Rombach, Blattmann, Lorenz, Esser, and Ommer]{rombach2022high}
Robin Rombach, Andreas Blattmann, Dominik Lorenz, Patrick Esser, and Bj{\"o}rn Ommer.
\newblock High-resolution image synthesis with latent diffusion models.
\newblock In \emph{Proceedings of the IEEE/CVF conference on computer vision and pattern recognition}, pages 10684--10695, 2022.

\bibitem[Shao et~al.(2024)Shao, Yang, Zhou, Zhang, Shen, Poggi, and Liao]{shao2024learning}
Jiahao Shao, Yuanbo Yang, Hongyu Zhou, Youmin Zhang, Yujun Shen, Matteo Poggi, and Yiyi Liao.
\newblock Learning temporally consistent video depth from video diffusion priors.
\newblock \emph{arXiv preprint arXiv:2406.01493}, 2024.

\bibitem[Shi et~al.(2024)Shi, Huang, Wang, Bian, Li, Zhang, Zhang, Cheung, See, Qin, et~al.]{shi2024motion}
Xiaoyu Shi, Zhaoyang Huang, Fu-Yun Wang, Weikang Bian, Dasong Li, Yi Zhang, Manyuan Zhang, Ka~Chun Cheung, Simon See, Hongwei Qin, et~al.
\newblock Motion-i2v: Consistent and controllable image-to-video generation with explicit motion modeling.
\newblock In \emph{ACM SIGGRAPH 2024 Conference Papers}, pages 1--11, 2024.

\bibitem[Teed and Deng(2020)]{teed2020raft}
Zachary Teed and Jia Deng.
\newblock Raft: Recurrent all-pairs field transforms for optical flow.
\newblock In \emph{European conference on computer vision}, pages 402--419. Springer, 2020.

\bibitem[Teed and Deng(2021)]{teed2021raft}
Zachary Teed and Jia Deng.
\newblock Raft-3d: Scene flow using rigid-motion embeddings.
\newblock In \emph{Proceedings of the IEEE/CVF conference on computer vision and pattern recognition}, pages 8375--8384, 2021.

\bibitem[Van~Hoorick et~al.(2022)Van~Hoorick, Tendulkar, Sur{\'\i}s, Park, Stent, and Vondrick]{van2022revealing}
Basile Van~Hoorick, Purva Tendulkar, D{\'\i}dac Sur{\'\i}s, Dennis Park, Simon Stent, and Carl Vondrick.
\newblock Revealing occlusions with 4d neural fields.
\newblock In \emph{Proceedings of the IEEE/CVF Conference on Computer Vision and Pattern Recognition}, pages 3011--3021, 2022.

\bibitem[Van~Hoorick et~al.(2024)Van~Hoorick, Wu, Ozguroglu, Sargent, Liu, Tokmakov, Dave, Zheng, and Vondrick]{vanhoorick2024gcd}
Basile Van~Hoorick, Rundi Wu, Ege Ozguroglu, Kyle Sargent, Ruoshi Liu, Pavel Tokmakov, Achal Dave, Changxi Zheng, and Carl Vondrick.
\newblock Generative camera dolly: Extreme monocular dynamic novel view synthesis.
\newblock 2024.

\bibitem[Voleti et~al.(2025)Voleti, Yao, Boss, Letts, Pankratz, Tochilkin, Laforte, Rombach, and Jampani]{voleti2025sv3d}
Vikram Voleti, Chun-Han Yao, Mark Boss, Adam Letts, David Pankratz, Dmitry Tochilkin, Christian Laforte, Robin Rombach, and Varun Jampani.
\newblock Sv3d: Novel multi-view synthesis and 3d generation from a single image using latent video diffusion.
\newblock In \emph{European Conference on Computer Vision}, pages 439--457. Springer, 2025.

\bibitem[Wang et~al.(2024{\natexlab{a}})Wang, Yang, Huang, Wang, and Li]{wang2024rectified}
Fu-Yun Wang, Ling Yang, Zhaoyang Huang, Mengdi Wang, and Hongsheng Li.
\newblock Rectified diffusion: Straightness is not your need in rectified flow.
\newblock \emph{arXiv preprint arXiv:2410.07303}, 2024{\natexlab{a}}.

\bibitem[Wang et~al.(2023)Wang, Chang, Cai, Li, Hariharan, Holynski, and Snavely]{wang2023omnimotion}
Qianqian Wang, Yen-Yu Chang, Ruojin Cai, Zhengqi Li, Bharath Hariharan, Aleksander Holynski, and Noah Snavely.
\newblock Tracking everything everywhere all at once.
\newblock In \emph{International Conference on Computer Vision}, 2023.

\bibitem[Wang et~al.(2024{\natexlab{b}})Wang, Ye, Gao, Austin, Li, and Kanazawa]{wang2024shape}
Qianqian Wang, Vickie Ye, Hang Gao, Jake Austin, Zhengqi Li, and Angjoo Kanazawa.
\newblock Shape of motion: 4d reconstruction from a single video.
\newblock \emph{arXiv preprint arXiv:2407.13764}, 2024{\natexlab{b}}.

\bibitem[Wang et~al.(2024{\natexlab{c}})Wang, Yuan, Wang, Li, Chen, Xia, Luo, and Shan]{wang2024motionctrl}
Zhouxia Wang, Ziyang Yuan, Xintao Wang, Yaowei Li, Tianshui Chen, Menghan Xia, Ping Luo, and Ying Shan.
\newblock Motionctrl: A unified and flexible motion controller for video generation.
\newblock In \emph{ACM SIGGRAPH 2024 Conference Papers}, pages 1--11, 2024{\natexlab{c}}.

\bibitem[Wu et~al.(2024)Wu, Yi, Fang, Xie, Zhang, Wei, Liu, Tian, and Wang]{wu20244d}
Guanjun Wu, Taoran Yi, Jiemin Fang, Lingxi Xie, Xiaopeng Zhang, Wei Wei, Wenyu Liu, Qi Tian, and Xinggang Wang.
\newblock 4d gaussian splatting for real-time dynamic scene rendering.
\newblock In \emph{Proceedings of the IEEE/CVF Conference on Computer Vision and Pattern Recognition}, pages 20310--20320, 2024.

\bibitem[Xiao et~al.(2024)Xiao, Wang, Zhang, Xue, Peng, Shen, and Zhou]{SpatialTracker}
Yuxi Xiao, Qianqian Wang, Shangzhan Zhang, Nan Xue, Sida Peng, Yujun Shen, and Xiaowei Zhou.
\newblock Spatialtracker: Tracking any 2d pixels in 3d space.
\newblock In \emph{Proceedings of the IEEE/CVF Conference on Computer Vision and Pattern Recognition (CVPR)}, 2024.

\bibitem[Yang et~al.(2024{\natexlab{a}})Yang, Kang, Huang, Zhao, Xu, Feng, and Zhao]{depth_anything_v2}
Lihe Yang, Bingyi Kang, Zilong Huang, Zhen Zhao, Xiaogang Xu, Jiashi Feng, and Hengshuang Zhao.
\newblock Depth anything v2.
\newblock \emph{arXiv:2406.09414}, 2024{\natexlab{a}}.

\bibitem[Yang et~al.(2024{\natexlab{b}})Yang, Teng, Zheng, Ding, Huang, Xu, Yang, Hong, Zhang, Feng, et~al.]{yang2024cogvideox}
Zhuoyi Yang, Jiayan Teng, Wendi Zheng, Ming Ding, Shiyu Huang, Jiazheng Xu, Yuanming Yang, Wenyi Hong, Xiaohan Zhang, Guanyu Feng, et~al.
\newblock Cogvideox: Text-to-video diffusion models with an expert transformer.
\newblock \emph{arXiv preprint arXiv:2408.06072}, 2024{\natexlab{b}}.

\bibitem[Yu et~al.(2023)Yu, Lezama, Gundavarapu, Versari, Sohn, Minnen, Cheng, Birodkar, Gupta, Gu, et~al.]{yu2023language}
Lijun Yu, Jos{\'e} Lezama, Nitesh~B Gundavarapu, Luca Versari, Kihyuk Sohn, David Minnen, Yong Cheng, Vighnesh Birodkar, Agrim Gupta, Xiuye Gu, et~al.
\newblock Language model beats diffusion--tokenizer is key to visual generation.
\newblock \emph{arXiv preprint arXiv:2310.05737}, 2023.

\bibitem[Zhang et~al.(2024)Zhang, Herrmann, Hur, Jampani, Darrell, Cole, Sun, and Yang]{zhang2024monst3r}
Junyi Zhang, Charles Herrmann, Junhwa Hur, Varun Jampani, Trevor Darrell, Forrester Cole, Deqing Sun, and Ming-Hsuan Yang.
\newblock Monst3r: A simple approach for estimating geometry in the presence of motion.
\newblock \emph{arXiv preprint arxiv:2410.03825}, 2024.

\end{thebibliography}
